%% file: main.tex
\documentclass[11pt]{article}
\usepackage[margin=1in]{geometry}
\usepackage[T1]{fontenc}
\usepackage{lmodern}
\usepackage{graphicx}
\usepackage{float}
\usepackage{booktabs}
\usepackage{xcolor}
\usepackage[numbers]{natbib}
\usepackage[hidelinks]{hyperref}
\usepackage{amsmath,amssymb}
\usepackage{booktabs}
\usepackage{xcolor}
\usepackage{tikz}
\usetikzlibrary{arrows.meta,decorations.pathreplacing}
\usepackage{pgfplots}
\pgfplotsset{compat=1.17}
\usepgfplotslibrary{groupplots}
\usepackage{colortbl}
\usepackage[numbers]{natbib}   
\input{numbers}



\title{\bf A Closer Look at Language Model Introspection: Probing Information and Reporting}
\author{Dustin Rubin, \and Vincenzo Timmel, \and Xufeng He, \and Aritra Saha, \and JooMan Kim, \and Vijay Madisetti \\ \\
\bf{Georgia Institute of Technology, Atlanta, GA 30332, USA}}
\date{November 10, 2026}

\begin{document}
\maketitle

\input{sections/abstract}
\input{sections/intro}

\input{sections/related}

\input{sections/method}
\input{sections/results}
\input{sections/discussion}
\input{sections/limitations}

\bibliographystyle{plainnat}
\bibliography{refs}

\clearpage
\appendix
\input{sections/appendix}

\end{document}

%% file: numbers.tex
\newcommand{\npFreshLo}{20.6}
\newcommand{\npFreshHi}{28.9}
\newcommand{\npUnlabelledLo}{11.1}
\newcommand{\npUnlabelledHi}{12.2}

\newcommand{\swProbeOneZeroThirtyTwo}{18.3}
\newcommand{\swProbeOneEarlyThirtyTwo}{76.9}

\newcommand{\swBaseAnswerThirtyTwo}{89.6}

\newcommand{\swBaseFixedAnswer}{37.7}

\newcommand{\swBaseFixedProbe}{83.1}

%% file: sections/abstract.tex
\begin{abstract}
One way to study language-model introspection is to test whether models can identify where their activations have been perturbed.
We train a simple linear classifier to identify the perturbed sentence from the model's activations, then compare its predictions with the model's own answers across pretraining and injection strengths. This tests whether the model is being asked to report information that is already easy to recover. If so, successful localization may be less demanding than it first appears and may not extend to the broader uses that motivate introspection research, such as faithfully explaining a model's reasoning or helping researchers assess the reliability of its answers.
With an injection four times the typical activation norm at layer 0 and probe readout at layer 24 in OLMo~3 32B, the supervised probe reaches 98.3\% accuracy after 8.4 billion training tokens, while the model's zero-shot answer remains at chance (11.1\%). At the released base checkpoint, these accuracies reach 99.1\% and 89.6\%, respectively.
Across 7B and 32B, location is already decodable in randomly initialized networks, showing that the architecture can support these traces without pretraining.
In these experiments, location information lies close enough to the surface for a simple linear probe to recover it, so successful reporting need not imply broader access to the model's internal computation.
\end{abstract}

%% file: sections/intro.tex
\section{Introduction}

Interest in language-model introspection emerges from the potential that large language models (LLMs) could faithfully report aspects of their internal computation, helping us understand and monitor their behavior \citep{lindsey2025emergent,hahami2025disturbance}.
Activation-perturbation experiments offer a controlled setting for studying these reports. Researchers change a model's internal activations while holding its input text fixed, then ask the model to detect or locate the change.
Successful perturbation localization shows that a model can report something about its internal computation, but its significance depends on what the task requires. A readily decodable trace could support successful localization without requiring a general ability to inspect internal computation. We therefore examine how accessible location information is to a simple linear probe before and after pretraining, and how this relates to the model's own reports across training stages.

\subsection{Sentence localization and reporting}
\citet{hahami2025disturbance} show that, in their experiments, injections can bias models toward affirmative answers, making yes/no detection misleading. Sentence localization requires the model to identify which sentence was altered, rather than simply report that a change occurred. This avoids the specific yes/no response bias, although preferences among sentence labels can still affect answers.
In our prompted task, a model reads nine numbered sentences, receives an activation perturbation at one sentence, and answers which sentence was altered. On the same trials, an external linear probe, a separately trained classifier, predicts the altered sentence from the model's activation at the position where it will answer.

\subsection{Our approach and main findings}
We examine whether perturbation-location information lies close enough to the surface of the model's representations for a simple linear probe to recover it, and what this implies for interpreting successful localization as evidence of introspection.

To examine how perturbation-location information develops during training, we use OLMo's publicly available pretraining checkpoints and released base models \citep{olmo3}. Our main comparison holds the injection and probe setting fixed across 32B training checkpoints.
At a fixed setting in 32B---injection layer 0, strength 4 and probe readout layer 24---the probe reaches 98.3\% accuracy after 8.4 billion training tokens while the model answers at chance. By the released base checkpoint, the probe reaches 99.1\% and the model reaches 89.6\%. The broader sweep tests how these results depend on injection layer, strength and model size.

The gap between decoding and reporting is highly sensitive to where the perturbation is introduced. At the released 32B base model, raw reporting accuracy was only 37.7\% when injected at layer 6, but climbed to 89.6\% when injected at layer 0 (both at strength $\alpha$=4). This demonstrates that a weak reporting score at one injection site does not imply a general inability of the model to localize perturbations and the result is highly configuration-dependent.

Location information is also decodable in networks with random, untrained weights. In the no-prompt experiment, strong random-direction injections leave traces that a trained linear probe can recover in three independently initialized networks, without setup instructions or a localization question. Decoding remains above chance at uninjected readouts after 256 additional tokens following the sentence list. Under these conditions, the architecture and random weights can propagate recoverable location information without pretraining.

%% file: sections/related.tex
\section{Related work}

\subsection{Introspective awareness through activation interventions}
Recent studies have investigated whether language models can recognize controlled interventions in their internal activations. \citet{lindsey2025emergent} examined introspective awareness through concept injection, while \citet{hahami2025disturbance} introduced sentence-localization tasks in which models identify which sentence in a multi-sentence input was affected by an activation perturbation. Unlike binary detection tasks, which ask whether a perturbation occurred, sentence localization requires models to identify the affected position among multiple candidates. This task can reduce some confounding effects of binary detection, such as reliance on general differences between perturbed and unperturbed inputs, and provides a more specific measure of perturbation-location reporting. \citet{macar2026mechanisms} further investigated the mechanisms of introspective detection and identification, showing that these capabilities depend on intervention conditions and model layers. \citet{lederman2026contentagnostic} further distinguish detecting an internal disturbance from identifying its content, finding that models can succeed at the former without reliably achieving the latter. Together, these studies establish activation interventions as a useful approach for evaluating functional introspection, while highlighting that introspective performance depends on the task and intervention procedure.

\subsection{Introspective capabilities across model training}
Recent work has examined how introspective capabilities change with model training. \citet{hahami2026introspectionfinetuningifttraining} demonstrated that sentence-localization performance can improve through targeted fine-tuning, while \citet{macar2026mechanisms} examined the influence of post-training on introspective performance. More directly, \citet{fornasiere2026perturbations} investigated localization of dropout and Gaussian-noise perturbations across OLMo pretraining checkpoints, finding that behavioral localization emerges during pretraining. Previous studies have also examined how introspective performance varies with intervention depth, defined as the transformer layer where a perturbation is introduced \citep{hahami2025disturbance,macar2026mechanisms}. However, existing developmental analyses have primarily focused on behavioral performance, leaving unclear how internally decodable perturbation-location information develops relative to explicit sentence-localization behavior across pretraining checkpoints.

\subsection{Internal representations and behavioral reporting}
Information encoded in internal representations does not necessarily translate into accurate behavioral reporting. In sentence-localization tasks, an external linear probe can be trained to predict the affected sentence from hidden activations, whereas behavioral localization evaluates the model's own sentence-number predictions. \citet{ferrara2026masked} found that intervention-related information could be recovered through linear probes even when models' verbal reports were unreliable.

\citet{singh2026reality} show that some introspection tasks can be solved using input cues or generic anomaly detection. \citet{tang2026causalbypass} show that accurate reports of internal signals can rely on correlated information rather than the particular signal being reported, and that training on interventions reduces this dependence.

How decodable location information develops relative to model reporting remains unclear. In particular, it is unclear whether these measurements follow similar trajectories during pretraining and how intervention conditions affect their measured differences. We address these questions by comparing linear-probe decodability and behavioral sentence localization across training, with additional tests of injection strength and randomly initialized networks.

%% file: sections/method.tex

\section{Method}
\label{sec:method}

We use two experiments: a pretraining sweep comparing a linear probe with model answers, and a no-prompt experiment testing propagation in random networks.

\subsection{Models}
\label{sec:models}

We use OLMo~3 \citep{olmo3} because the weights of every training stage are public.
Unless stated otherwise we use the 32B model (64 blocks, width $d = 5120$).
We compare first-stage pretraining checkpoints and the released base models.
Models run in bf16 with PyTorch forward hooks in Hugging Face Transformers.

\subsection{The localization test}
\label{sec:task}

We use a nine-sentence version of the localization test of \citet{hahami2025disturbance}.
Each prompt has a fixed setup text, nine numbered sentences, a question and the start of the answer
(Figure~\ref{fig:task}a).
The nine sentences are drawn without replacement from a pool of 100 short, neutral sentences such as
``The dog barks loudly.''

The sweep prompt ends with \texttt{Answer: SENTENCE\textvisiblespace}.
We run one forward pass and select the highest-scoring label among 1--9 from the model's logits at the final input position $T$, taking the maximum over token variants for each label (chance of selecting the correct sentence: $1/9$).

\subsection{Injection}
\label{sec:injection}

\begin{figure}[H]
\centering
\begin{minipage}[b]{0.53\linewidth}
\centering
\fcolorbox{black!30}{black!3}{%
\parbox{0.95\linewidth}{\footnotesize\ttfamily\raggedright
Below are numbered sentences. While the model reads them, one or more sentences may be altered from the inside:
not in their words, but in the model's internal processing of them.\\[4pt]
SENTENCE 1: The cat sleeps on the couch.\\
SENTENCE 2: She reads a book quietly.\\
SENTENCE 3: The train arrives at noon.\\
SENTENCE 4: \colorbox{orange!40}{Rain falls on the roof.}\\
SENTENCE 5: The child plays in the park.\\
SENTENCE 6: He drinks his morning coffee.\\
SENTENCE 7: A bird flies over the lake.\\
SENTENCE 8: The teacher writes on the board.\\
SENTENCE 9: The dog barks loudly.\\[4pt]
Which sentence was altered?\\
Answer: SENTENCE\colorbox{blue!15}{\textvisiblespace}\textrm{\textit{[score next digit]}}}}
\par\smallskip (a) Prompt
\end{minipage}\hfill
\begin{minipage}[b]{0.44\linewidth}
\centering
\begin{tikzpicture}[x=0.42cm, y=0.38cm, font=\scriptsize, >={Stealth[length=4pt]}]
  \foreach \x in {0,...,10} {
    \foreach \y in {0,...,7} {
      \def\cellcol{black!8}
      \ifnum\y>2 \ifnum\x>2 \def\cellcol{orange!25}\fi\fi
      \ifnum\y=2 \ifnum\x>2 \ifnum\x<6 \def\cellcol{orange!85}\fi\fi\fi
      \fill[\cellcol] (\x+0.07,\y+0.08) rectangle (\x+0.93,\y+0.92);
    }
  }
  \node[white] at (4.5,2.5) {$+\alpha r_L \hat{v}$};
  \draw[->, thick, orange!60!black] (5.7,2.7) .. controls (7.5,2.9) and (9.0,3.1) .. (10.2,3.45);
  \draw[blue!70!black, very thick] (10.03,5.05) rectangle (10.97,5.95);
  \draw[->, thick, blue!70!black] (11.1,5.5) -- (11.9,5.5) node[right] {probe};
  \draw[->, thick] (10.5,8.0) -- (10.5,8.8) node[above] {$s_1, \dots, s_9$};
  \node[anchor=east] at (-0.1,0.5) {$0$};
  \node[anchor=east] at (-0.1,2.5) {$L$};
  \node[anchor=east] at (-0.1,5.5) {$\ell$};
  \node[anchor=east] at (-0.1,7.5) {$N{-}1$};
  \node[rotate=90] at (-1.5,4) {block};
  \node at (1,-0.6) {$S_1$};
  \node at (2.5,-0.6) {$\cdots$};
  \node at (4.5,-0.6) {$S_k$};
  \node at (7,-0.6) {$\cdots$};
  \node at (9,-0.6) {$Q$};
  \node[fill=blue!15, inner sep=2pt] at (10.5,-0.6) {$T$};
  \node at (5.5,-1.6) {token};
\end{tikzpicture}
\par\smallskip (b) Residual stream
\end{minipage}
\caption{The localization test.
(a) A sweep prompt with sentence 4 injected. The shaded trailing space is the final input token; its logits score the next digit.
(b) The residual stream, one column per token and one row per block; $Q$ marks the question tokens and $T$ the last token (blue, corresponding to the trailing space in (a)).
Dark orange: the injected tokens of $S_k$ at block $L$.
Light orange: cells that can be affected through later attention to the injected tokens.
Gray: unchanged.
The blue box is the probe's readout, at $T$ in block $\ell$.}
\label{fig:task}
\end{figure}
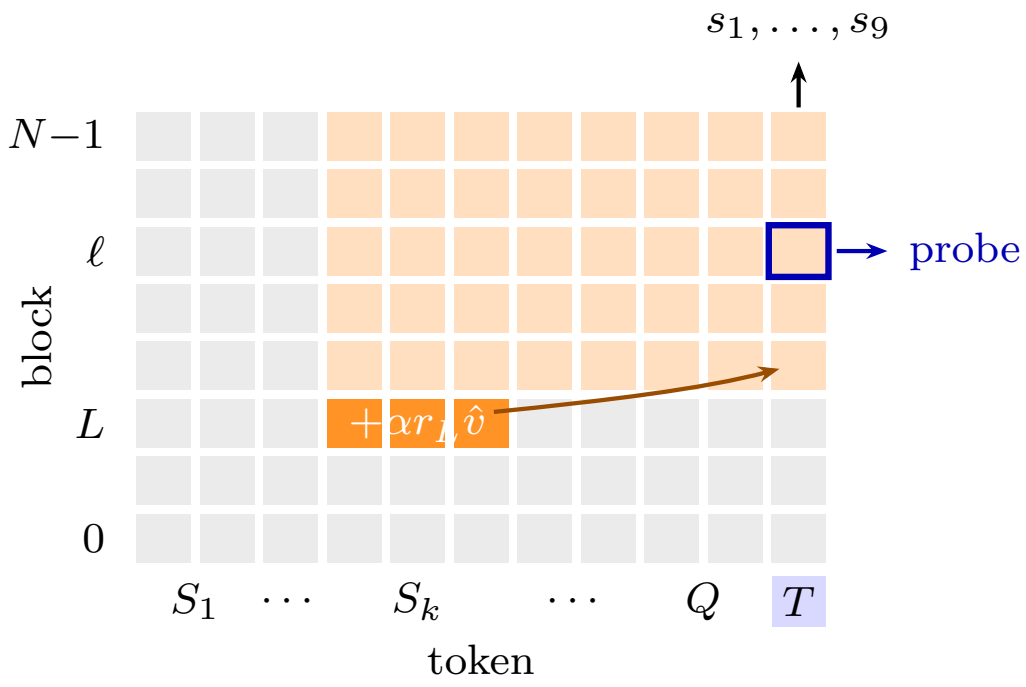

Let $h_t^{(\ell)} \in \mathbb{R}^d$ be the residual stream at token $t$ after decoder block $\ell$ (the hidden state
passed from block $\ell$ to block $\ell{+}1$), with blocks numbered from $0$ to $N{-}1$ ($N = 64$ for OLMo~3 32B).
To inject into sentence $k$ at layer $L$, a forward hook adds the same vector to every token of that sentence:
\begin{equation}
  h_t^{(L)} \;\leftarrow\; h_t^{(L)} + \alpha \, r_L \, \hat{v}
  \qquad \text{for all } t \in S_k ,
  \label{eq:inject}
\end{equation}
where $S_k$ holds the token positions of the sentence's words (not of ``\texttt{SENTENCE $k$:}''), $\hat{v}$ is a
unit-length direction, $r_L$ is the typical residual norm at layer $L$, and $\alpha$ is the injection strength
(Section~\ref{sec:strength}).
After injection, attention in subsequent blocks can carry the perturbation to later tokens (Figure~\ref{fig:task}b).

We use two types of injected vector $\hat{v}$.
The prompted comparisons pool concept-vector and random-direction trials equally. The no-prompt experiment uses random directions only.

\subsubsection{Concept vectors}
Concept vectors are computed separately for each checkpoint and layer.
For a concept word $c$, $\hat{v}$ is the mean of $h^{(L)}$ over the tokens of $c$ in ``\texttt{Here is a word: $c$}'',
minus the same mean averaged over 99 baseline words, normalized to unit length.
Each concept-vector passage draws one of 19 concepts at random.

\subsubsection{Random vectors}
\[z \sim \mathcal{N}(0, I_d) \quad \text{and} \quad \hat{v} = z / \lVert z \rVert.\]

\subsubsection{Strength}
\label{sec:strength}
We set the size of the injected vector relative to the model's own activations.
For the prompted experiments, we take one clean prompt per checkpoint (the first nine sentences of the pool) and compute the median norm of its sentence-token residuals at layer $L$:
\begin{equation}
  r_L = \operatorname*{median}_{t} \, \lVert h_t^{(L)} \rVert_2 ,
\end{equation}
over the tokens $t$ of the nine sentences. The no-prompt experiment instead calibrates on nine sentences from its training pool, without setup text or a question.
Because $\hat{v}$ has unit length, the vector added in Equation~\ref{eq:inject} has norm $\alpha\, r_L$.
So $\alpha = 1$ adds a vector as long as a typical token's residual, and $\alpha = 4$ one four times as long.

\subsection{Measurements}
\label{sec:measures}

\paragraph{Linear probe.}
A nine-way logistic regression predicts the injected sentence from the residual $h_T^{(\ell)}$ at the last prompt token, at layers after the injection ($\ell > L$).
The language model's weights remain fixed while the probe is trained. Probe accuracy is measured on held-out trials, using passage-grouped five-fold cross-validation for the pretraining sweep and separate training and test sentence pools for the no-prompt experiment.

\paragraph{Model's answer.}
The highest-scoring label among 1--9 at the last prompt token (Section~\ref{sec:task}).
We call it the raw answer.

\subsection{Experimental settings}
\label{sec:settings}

\subsubsection{Training stages}
We compare seven first-stage pretraining checkpoints and the released base model, which includes subsequent mid-training and long-context training. After examining a wider layer and strength sweep in OLMo~3 7B and 32B, we focus on a fixed setting in 32B: injection at layer 0, strength $\alpha=4$ and probe readout at layer 24. This keeps the setting constant across the training trajectory and uses the injection layer and strength selected for the released base model's highest raw answer accuracy in all three held-out-seed folds.

\subsubsection{No-prompt localization}
We inject random directions at layer 6 into nine numbered sentences, without setup instructions or a question, and read layer 24 at uninjected markers after the list.
We test three randomly initialized networks across injection strengths and readout distances.
Probes use disjoint training and test sentence pools.

%% file: sections/results.tex
\section{Results}

Reported ranges are across seeds. Each sweep probe is trained on 288 labeled trials per fold and evaluated on held-out passages. The model answers without labeled examples in its prompt.

\subsection{Decoding and reporting across pretraining}
\label{sec:sweep}

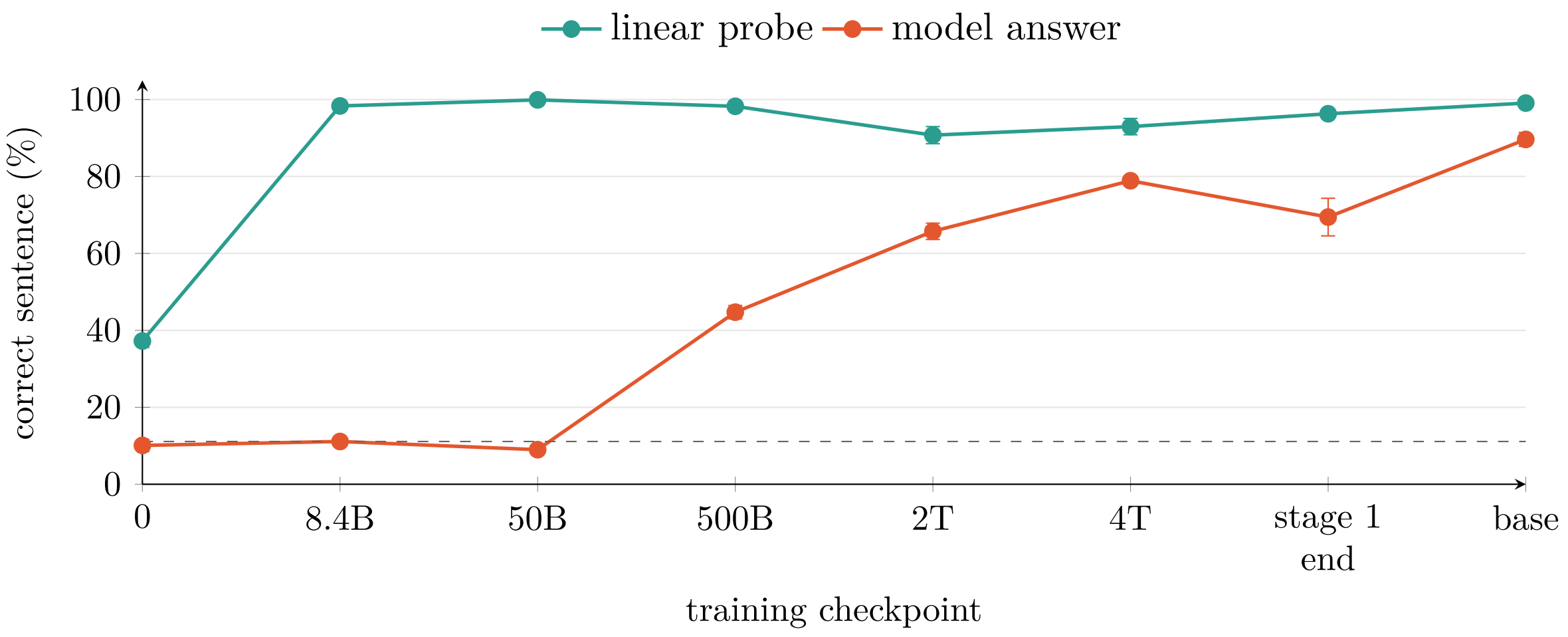
\begin{figure}[t]
  \centering
  \input{figures/fig_fixed_pretraining}
  \caption{OLMo 3 32B across pretraining, with injection at layer 0, strength $\alpha=4$ and probe readout at layer 24 throughout. This fixed-setting analysis was chosen after examining the sweep. Lines show means across three data seeds; bars show one standard deviation. Concept and random directions are pooled. The dashed line marks one-in-nine chance. B and T denote billions and trillions of training tokens; positions are categorical, and the released base includes subsequent training.}
  \label{fig:fixed-pretraining}
\end{figure}

With injection layer, strength and probe readout held fixed, location information becomes easy to decode early in pretraining, while the model's answers improve later (Figure~\ref{fig:fixed-pretraining}). In 32B, the probe reaches 98.3\% after 8.4 billion training tokens, while the model answers at 11.1\%, equal to chance. At the released base checkpoint, the probe reaches 99.1\% and the model reaches 89.6\%. Successful reporting therefore develops without post-training, but follows the early increase in probe accuracy.

Location information is already decodable before pretraining: at step 0, the same probe setting gives 37.2\% accuracy, compared with 11.3\% for shuffled labels. At the lower strength $\alpha=2$, step-0 decoding reaches 30.5\%, compared with 10.1\% for shuffled labels. This comparison selects the probe readout on two seeds and evaluates it on the third; all three folds select layer 60, rather than the main trajectory's fixed layer 24. The gap between probe and answer depends on the injection setting. The broader sweep across both model sizes, including lower strengths and the layer-6 comparison, is reported in Appendix~\ref{app:sweep-results}.

\subsection{Propagation without pretraining or a question}
\label{sec:noprompt}

The step-0 result leaves open whether recoverable location information depends on the localization prompt. The no-prompt experiment tests whether it also reaches later tokens without setup instructions or a localization question. In three independently initialized random networks, a probe at an uninjected token identifies the perturbed sentence above chance at the larger injection strengths. At strength 8, accuracy is 31.5\% immediately after the list and 21.5\% after 256 additional tokens, compared with uninjected controls of 11.5\% and 10.9\%. The signal is therefore recoverable without pretraining or a localization question, including at readouts separated from the list by additional text.

This result is not confined to injections in the final sentence: accuracy for targets in sentences 1--8 is 30.6\% at gap 0 and 21.9\% at gap 256, using the same nine-choice decoder. The pretrained comparison has an unresolved elevated uninjected control, so we do not use it to draw conclusions about changes in propagation during training.

%% file: figures/fig_fixed_pretraining.tex
\definecolor{fixedprobe}{HTML}{2A9D8F}
\definecolor{fixedanswer}{HTML}{E4572E}
\begin{tikzpicture}
\begin{axis}[
 width=0.93\linewidth, height=5.6cm,
 symbolic x coords={0,8.4B,50B,500B,2T,4T,end,base},
 xtick=data, xticklabels={0,8.4B,50B,500B,2T,4T,{stage 1\\end},base},
 xticklabel style={align=center}, enlarge x limits=0.05,
 ymin=0,ymax=105,ytick={0,20,...,100},
 xlabel={training checkpoint},ylabel={correct sentence (\%)},
 legend style={at={(0.5,1.05)},anchor=south,draw=none},legend columns=2,
 ymajorgrids,grid style={black!10},axis lines=left,
 tick label style={font=\small},label style={font=\small}]
\addplot[fixedprobe,line width=1pt,mark=*,error bars/.cd,y dir=both,y explicit] coordinates {(0,37.2222) +- (0,1.6667) (8.4B,98.3333) +- (0,0.4811) (50B,99.9074) +- (0,0.1604) (500B,98.2407) +- (0,1.2526) (2T,90.7407) +- (0,2.2280) (4T,92.9630) +- (0,2.1216) (end,96.2963) +- (0,0.8929) (base,99.0741) +- (0,0.1604)};
\addlegendentry{linear probe}
\addplot[fixedanswer,line width=1pt,mark=*,error bars/.cd,y dir=both,y explicit] coordinates {(0,10.0926) +- (0,1.5795) (8.4B,11.1111) +- (0,0.0000) (50B,8.9815) +- (0,0.6991) (500B,44.7222) +- (0,1.7347) (2T,65.7407) +- (0,2.1216) (4T,78.8889) +- (0,0.2778) (end,69.4444) +- (0,4.8829) (base,89.6296) +- (0,1.8073)};
\addlegendentry{model answer}
\addplot[black!60,dashed,forget plot] coordinates {(0,11.1111) (base,11.1111)};
\end{axis}\end{tikzpicture}

%% file: sections/discussion.tex
\section{Discussion}

\subsection{Information and reporting across training}
Location information becomes highly decodable early in pretraining, while reliable reporting develops later. Under suitable injection settings, the base model can report the location without post-training. The apparent gap between decoding and reporting depends strongly on where the perturbation is introduced: at strength 4 in the released 32B base, raw reporting reaches 37.7\% with injection at layer 6 and 89.6\% at layer 0. A weak answer score at one injection site therefore does not establish a general inability to localize. Comparisons across training stages need to distinguish changes in the model from changes in the conditions under which it is tested.
Our comparison adds a measure of decodable information to the behavioral trajectories studied by \citet{fornasiere2026perturbations}. It does not directly test the emergence of yes-or-no detection during preference optimization reported by \citet{macar2026mechanisms}, which uses a different task.

\subsection{Propagation before pretraining}
The sweep shows above-chance decoding at step 0. The no-prompt experiment extends this result: in independently initialized random networks, location information reaches later, uninjected tokens even without a localization question. At higher strengths, it remains recoverable after 256 additional tokens. Localization instructions are therefore not required for a recoverable location signal to form. A question may affect how the model uses the signal, but it is not necessary for the signal to exist in this setting.
These results show that the architecture and random weights can carry a location signal without pretraining. They do not establish how pretraining changes this propagation, because the pretrained no-prompt comparison has an unresolved uninjected-control anomaly.

\subsection{Implications for introspection claims}
The availability of a linearly decodable location signal, including before pretraining, provides a possible basis for successful reporting without requiring a general ability to inspect internal computation. \citet{singh2026reality} identify input cues and generic anomaly detection as alternative explanations for success on some introspection tasks; our experiments examine how readily an internal perturbation leaves a decodable location trace while the input text is held fixed, including before pretraining. \citet{tang2026causalbypass} show that accurate self-reports can rely on correlated signals rather than the signal being reported, underscoring why our probe--answer comparison does not establish that the model uses the features recovered by the probe. These findings therefore do not rule out introspection, but they limit what localization success alone can establish about a model's ability to report other aspects of its computation.

%% file: sections/limitations.tex
\section{Limitations}
An external linear probe can recover perturbation-location information even when the model's own reporting answer is incorrect. Successful decoding does not establish that the model bases its response on features decoded by the probe. Attention interventions help us understand how perturbations influence subsequent activations but do not establish a causal link between the decoded features and the model's response.

Artificial activation perturbations were introduced into OLMo models that processed short lists of sentences, which differs from reporting on computations arising during normal inference without perturbation. The no-prompt experiment tested three random initializations under one Gaussian scheme, leaving other schemes and architectures unexplored. Frozen probes were sensitive to modifications made in text and numbering. Decreased transfer does not indicate that information about location was absent. Removing sentence labels also shortens the input, so this test does not isolate the effect of numbering alone. Instead of evaluating unconstrained generation, the localization assessment scored nine candidate labels.

Scaling injections relative to residual norms does not ensure comparable functional disruption across checkpoints. The attention-path checks used only two held-out trials per target per seed, limiting their coverage of input variation. Some no-prompt reruns did not reproduce saved activations within the preset numerical tolerance; results depended on the attention implementation.
The elevated pretrained uninjected-control scores remain unexplained. Sweep exports provide aggregate accuracy rather than passage-level predictions, so the figures show variation across three seeds rather than bootstrap confidence intervals. The main fixed setting was chosen after inspecting the sweep. In the supporting selected-setting curves, the three held-out scores reuse seeds for setting selection, so their error bars should not be read as uncertainty from three independent experiments.

%% file: sections/appendix.tex
\section{Input templates}
\label{app:prompts}
Braced fields denote sampled sentences; they are not literal prompt text.
All prompted runs use plain text, without user/assistant chat-role tokens.

\par\bigskip\noindent\begin{minipage}{\linewidth}
\subsection{Prompted localization}
\begin{quote}
\small\ttfamily\raggedright
Below are numbered sentences. While the model reads them, one or more sentences may be altered from the inside: not in their words, but in the model's internal processing of them.\\*[\baselineskip]
SENTENCE 1: \{sentence 1\}\\*
SENTENCE 2: \{sentence 2\}\\*
SENTENCE 3: \{sentence 3\}\\*
SENTENCE 4: \{sentence 4\}\\*
SENTENCE 5: \{sentence 5\}\\*
SENTENCE 6: \{sentence 6\}\\*
SENTENCE 7: \{sentence 7\}\\*
SENTENCE 8: \{sentence 8\}\\*
SENTENCE 9: \{sentence 9\}\\*[\baselineskip]
Which sentence was altered?\\*
Answer: SENTENCE\textvisiblespace
\end{quote}
\end{minipage}\par\medskip

\textbf{Injection and evaluation.}
The injection applies to the word tokens of one target sentence, excluding its numbered prefix.
The model's answer is the highest-scoring label among 1--9 at the final input position; the external probe reads activations there.
The input ends with \texttt{SENTENCE\textvisiblespace}, including the trailing space and no answer digit.

\par\bigskip\noindent\begin{minipage}{\linewidth}
\subsection{No-prompt input}
\textbf{Sentence list}
\begin{quote}
\small\ttfamily\raggedright
SENTENCE 1: \{sentence 1\}\\*
SENTENCE 2: \{sentence 2\}\\*
SENTENCE 3: \{sentence 3\}\\*
SENTENCE 4: \{sentence 4\}\\*
SENTENCE 5: \{sentence 5\}\\*
SENTENCE 6: \{sentence 6\}\\*
SENTENCE 7: \{sentence 7\}\\*
SENTENCE 8: \{sentence 8\}\\*
SENTENCE 9: \{sentence 9\}
\end{quote}
\end{minipage}\par\medskip

\textbf{Suffix and readout.}
There is no setup, question, or answer prefix.
The list is tokenized first. A neutral suffix is built by tokenizing sampled sentences, each followed by a space, and truncating the resulting token sequence to 257 tokens.
Suffix tokens at zero-based offsets 0, 16, 64, and 256 are replaced with the single newline token.
These newline tokens are the readout positions; offset 0 is immediately after the sentence list.
The suffix is appended as token IDs, not re-tokenized as a combined string.

\textbf{Injection and evaluation.}
A random direction is added to the word tokens of one target sentence at layer 6.
The primary location probe reads layer 24 at the uninjected newline markers.
All markers follow every sentence's word tokens; the model is not asked to produce an answer.

\section{Concepts, checkpoints and evaluation}
\label{app:models}

\subsection{Concept vectors}
\label{app:concepts}
The 19 concepts are Dust, Satellites, Trumpets, Origami, Illusions, Volcanoes, Umbrellas, Violins, Lanterns, Cactus, Submarines, Pancakes, Chess, Lighthouses, Kites, Dinosaurs, Perfume, Bridges and Helicopters.

\subsection{Models}
The 32B released base is \path{allenai/Olmo-3-1125-32B}.
Stage-1 revisions are named \texttt{stage1-step} followed by the step number.
The 7B revisions belong to \path{allenai/Olmo-3-1025-7B}.
\begin{center}
\small
\begin{tabular}{lll}
\toprule
Training stage & 32B revision & 7B revision \\
\midrule
step 0 & step0 & step0 \\
about 8.4B tokens & step1000 & step2000 \\
about 50B tokens & step6000 & step12000 \\
about 500B tokens & step60000 & step119000 \\
about 2T tokens & step238000 & step476838 \\
about 4T tokens & step477000 & step954000 \\
end of stage 1 & step656000 & step1413814 \\
released base & \texttt{main} & \texttt{main} \\
\bottomrule
\end{tabular}
\end{center}
B and T denote billions and trillions of training tokens.

\subsection{Experiment settings}
\label{app:setup}
The sweep uses seeds 21, 101 and 102, with 40 passages per seed. Each passage is injected at all nine sentence positions, giving 360 trials per condition.
Injection strengths are $\alpha\in\{0.25,0.5,1,2,4\}$; injection layers are $\{0,6,16,32,48\}$ for 32B and $\{0,3,8,16,24\}$ for 7B.
Half the passages use concept vectors and half use random directions. Each passage retains its vector across target positions and strengths. Seeds vary passages and vectors, not the model's initial weights.

\subsection{Strength comparison with prior work}
\citet{hahami2025disturbance} report 88\% localization accuracy with coefficient 5 on a unit-normalized vector at layer 2 of Llama-3.1-8B-Instruct.
Our calibration on 40 neutral sentences measured a median residual norm of 2.464 at that block's output, giving a relative injection size of $5/2.464\approx2.03$, close to our $\alpha=2$.
This is an approximate comparison of relative magnitude across models and calibration text, not matched functional disruption.

\subsection{Linear probe}
\label{app:probe}
The sweep fits nine-way logistic regression with an L2 penalty, $C=0.5$, and a maximum of 3{,}000 iterations. Features are standardized using the training fold's mean and standard deviation.
Readouts are every fourth layer and the final layer, strictly after injection.
Five-fold evaluation groups trials by passage, keeping all nine target positions of a passage in the same fold. Sentences and concepts can recur across folds, so these scores do not test unseen sentence pools or concepts.
Each probe is trained on four folds and tested on the fifth; we report mean test accuracy. One frozen shuffled-label control is fitted per readout dataset.
For the reported step-0 comparison at $\alpha=2$ and injection layer 0, shuffled-label accuracy averages 10.1\% in 32B and 11.6\% in 7B at the readouts selected by the original probes. The controls are not separately selected to maximize their scores.
The separate no-prompt experiment uses disjoint training and test sentence pools (Appendix~\ref{app:noprompt-details}).

\input{sections/sweep_support}
\input{sections/results_details}

%% file: sections/sweep_support.tex
\section{Supporting sweep results}
\label{app:sweep-results}

\subsection{Layer and strength selection}
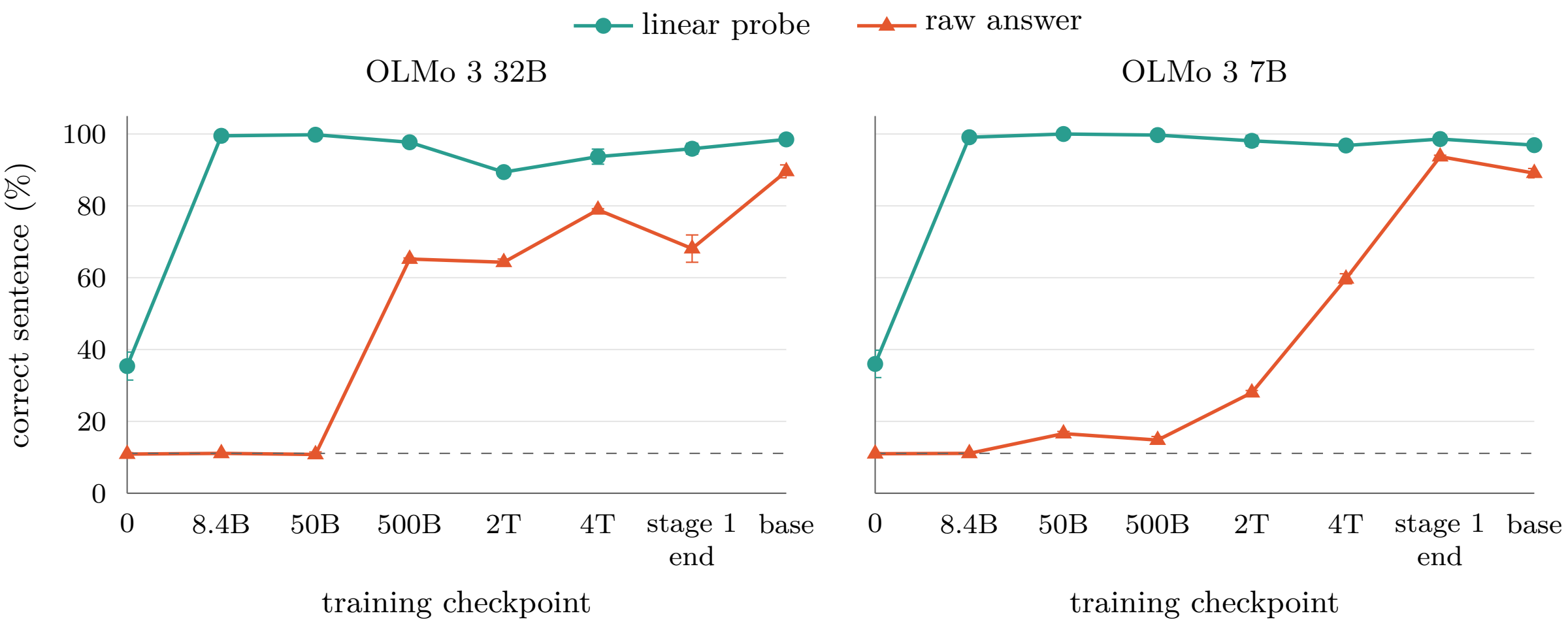
\begin{figure}[t]
  \centering
  \input{figures/fig_sweep}
  \caption{Probe and raw answer across stage-1 pretraining and the released base. Each measure independently selects injection layer and strength on two seeds and is scored on the third; the probe also selects readout layer. Lines average the three held-out scores; bars show one standard deviation across seeds, not confidence intervals. Concept and random directions are pooled. B and T denote billions and trillions of training tokens. Positions are categorical; the released base includes subsequent training. Chance is one in nine.}
  \label{fig:sweep}
\end{figure}

The full sweep covers 7B and 32B. The curves select injection layer and strength separately for each measure, and readout layer for the probe. Settings are selected by mean accuracy on two seeds and evaluated on the third, repeating for all three held-out seeds (21, 101 and 102). Figure~\ref{fig:sweep} averages these held-out scores. The probe and model answer may therefore use different settings. The main fixed-setting analysis was chosen after examining this sweep, rather than prespecified.

\subsection{Fixed layer-6 comparison and controls}
At a common setting in the released 32B base---injection layer 6, $\alpha=4$ and probe layer 24---the probe reaches \swBaseFixedProbe\% (80.8--85.6\% across seeds) while the model answers correctly on \swBaseFixedAnswer\% (35.8--41.1\%).
Raw reporting reaches \swBaseAnswerThirtyTwo\% when its injection setting is selected, showing that weak reporting at one setting is not a general inability to localize.
The shuffled-label probe scores 12.0\% (10.8--12.8\%). Evaluating the same pooled probe separately gives 86.1\% for concept injections and 80.2\% for random directions; corresponding raw answers are 42.8\% and 32.6\%.

\subsection{Strength dependence}
\begin{figure}[t]
  \centering
  \input{figures/fig_strength_step}
  \caption{Probe accuracy by injection strength at three checkpoints, with injection at layer 0. The readout layer is chosen on two seeds and scored on the third. Lines are means over three held-out seeds; bars show one standard deviation across seeds, not confidence intervals. Concept and random directions are pooled. Doses double between equally spaced ticks. Chance is one in nine.}
  \label{fig:strength-step}
\end{figure}
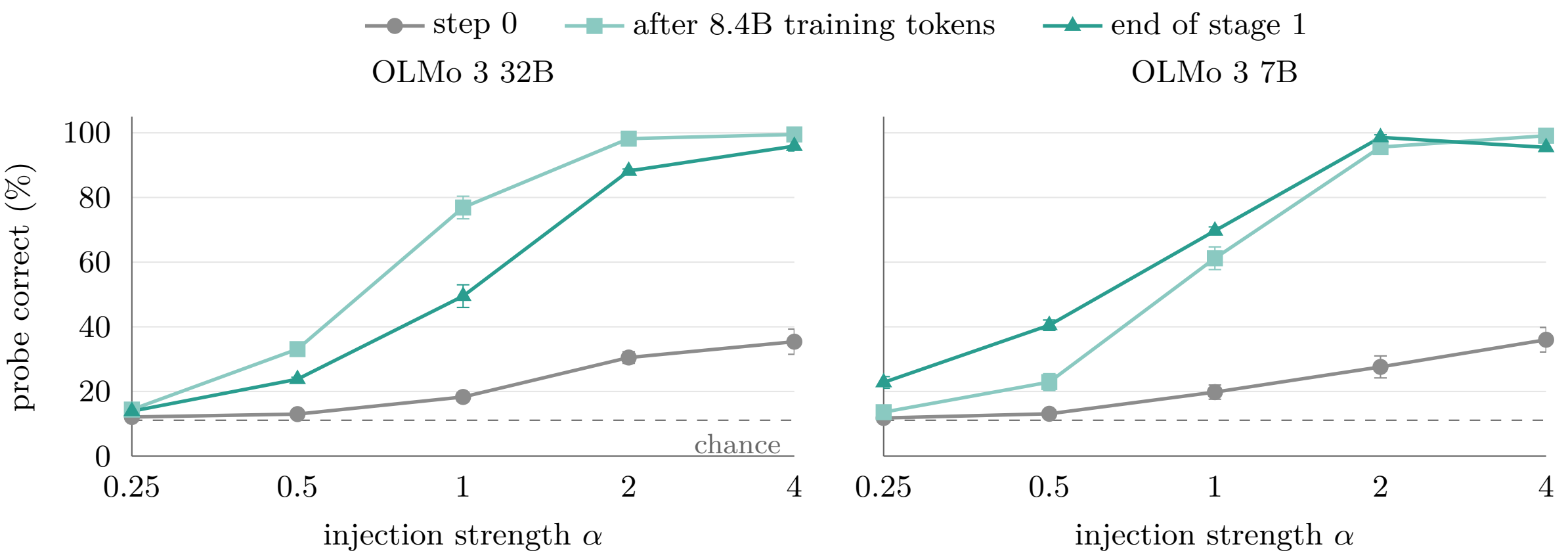

The early improvement in decoding is also present at a smaller injection strength (Figure~\ref{fig:strength-step}).
With injection at layer 0 and $\alpha=1$, 32B probe accuracy rises from \swProbeOneZeroThirtyTwo\% at step 0 to \swProbeOneEarlyThirtyTwo\% after 8.4 billion training tokens; 7B shows the same direction of change.
At $\alpha=2$ and injection layer 0, step-0 accuracy is 30.5\% in 32B (28.6--32.2\%) and 27.6\% in 7B (23.9--30.6\%), using the same held-out-seed readout selection.
The strength profiles also show that decoding does not improve uniformly throughout training: at some strengths, an early checkpoint is more decodable than the end of stage 1.

%% file: figures/fig_sweep.tex
\definecolor{probecolor}{HTML}{2A9D8F}
\definecolor{answercolor}{HTML}{E4572E}
\begin{tikzpicture}
\begin{groupplot}[
  group style={group size=2 by 1, horizontal sep=0.9cm, ylabels at=edge left, yticklabels at=edge left},
  width=0.5\linewidth, height=5.4cm,
  symbolic x coords={0,8.4B,50B,500B,2T,4T,end,base},
  xtick={0,8.4B,50B,500B,2T,4T,end,base}, xticklabels={0,8.4B,50B,500B,2T,4T,{stage 1\\end},base},
  xticklabel style={align=center}, enlarge x limits=0.06,
  ymin=0, ymax=105, ytick={0,20,...,100},
  xlabel={training checkpoint}, ylabel={correct sentence (\%)},
  legend columns=2, legend style={draw=none, font=\footnotesize, /tikz/every even column/.append style={column sep=1em}},
  ymajorgrids, grid style={black!10},
  axis lines=left, axis line style={-, black!60}, tick style={draw=none},
  tick label style={font=\scriptsize}, label style={font=\footnotesize}, title style={font=\footnotesize},
]
\nextgroupplot[title={OLMo 3 32B}, legend to name=sweeplegend]
\addplot[probecolor, line width=1pt, mark=*, mark size=1.8pt, error bars/.cd, y dir=both, y explicit] coordinates {(0,35.4) +- (0,3.9) (8.4B,99.5) +- (0,0.2) (50B,99.8) +- (0,0.2) (500B,97.7) +- (0,0.7) (2T,89.4) +- (0,1.1) (4T,93.7) +- (0,2.1) (end,95.9) +- (0,1.5) (base,98.5) +- (0,1.1)};
\addlegendentry{linear probe}
\addplot[answercolor, line width=1pt, mark=triangle*, mark size=1.8pt, error bars/.cd, y dir=both, y explicit] coordinates {(0,10.9) +- (0,0.4) (8.4B,11.1) +- (0,0.0) (50B,10.8) +- (0,0.7) (500B,65.2) +- (0,0.3) (2T,64.3) +- (0,0.9) (4T,78.9) +- (0,0.3) (end,68.1) +- (0,3.8) (base,89.6) +- (0,1.8)};
\addlegendentry{raw answer}
\addplot[black!60, dashed, forget plot] coordinates {(0,11.1) (base,11.1)};
\nextgroupplot[title={OLMo 3 7B}]
\addplot[probecolor, line width=1pt, mark=*, mark size=1.8pt, error bars/.cd, y dir=both, y explicit] coordinates {(0,36.0) +- (0,3.8) (8.4B,99.1) +- (0,0.6) (50B,100.0) +- (0,0.0) (500B,99.7) +- (0,0.3) (2T,98.1) +- (0,1.5) (4T,96.8) +- (0,0.2) (end,98.6) +- (0,0.8) (base,96.9) +- (0,0.2)};
\addplot[answercolor, line width=1pt, mark=triangle*, mark size=1.8pt, error bars/.cd, y dir=both, y explicit] coordinates {(0,11.0) +- (0,0.3) (8.4B,11.1) +- (0,0.0) (50B,16.6) +- (0,0.6) (500B,14.8) +- (0,1.0) (2T,28.0) +- (0,0.6) (4T,59.7) +- (0,1.4) (end,93.7) +- (0,0.4) (base,89.1) +- (0,1.3)};
\addplot[black!60, dashed, forget plot] coordinates {(0,11.1) (base,11.1)};
\end{groupplot}
\path (group c1r1.north west) -- (group c2r1.north east) node[midway, above=0.5cm] {\pgfplotslegendfromname{sweeplegend}};
\end{tikzpicture}

%% file: figures/fig_strength_step.tex
\definecolor{probecolor}{HTML}{2A9D8F}
\begin{tikzpicture}
\begin{groupplot}[
  group style={group size=2 by 1, horizontal sep=0.9cm, ylabels at=edge left, yticklabels at=edge left},
  width=0.5\linewidth, height=5cm,
  symbolic x coords={0.25,0.5,1,2,4}, xtick={0.25,0.5,1,2,4}, enlarge x limits=0.08,
  ymin=0, ymax=105, ytick={0,20,...,100},
  xlabel={injection strength $\alpha$}, ylabel={probe correct (\%)},
  legend columns=3, legend style={draw=none, font=\footnotesize, /tikz/every even column/.append style={column sep=1em}},
  ymajorgrids, grid style={black!10},
  axis lines=left, axis line style={-, black!60}, tick style={draw=none},
  tick label style={font=\footnotesize}, label style={font=\footnotesize}, title style={font=\footnotesize},
]
\nextgroupplot[title={OLMo 3 32B}, legend to name=strengthsteplegend]
\addplot[black!45, line width=1pt, mark=*, mark size=1.8pt, error bars/.cd, y dir=both, y explicit] coordinates {(0.25,12.1) +- (0,0.4) (0.5,13.0) +- (0,0.2) (1,18.3) +- (0,1.4) (2,30.5) +- (0,1.8) (4,35.4) +- (0,3.9)};
\addlegendentry{step 0}
\addplot[probecolor!55, line width=1pt, mark=square*, mark size=1.8pt, error bars/.cd, y dir=both, y explicit] coordinates {(0.25,14.4) +- (0,1.3) (0.5,33.1) +- (0,1.5) (1,76.9) +- (0,3.5) (2,98.2) +- (0,0.2) (4,99.5) +- (0,0.2)};
\addlegendentry{after 8.4B training tokens}
\addplot[probecolor, line width=1pt, mark=triangle*, mark size=1.8pt, error bars/.cd, y dir=both, y explicit] coordinates {(0.25,13.9) +- (0,0.8) (0.5,23.8) +- (0,0.6) (1,49.5) +- (0,3.5) (2,88.2) +- (0,0.6) (4,95.9) +- (0,1.5)};
\addlegendentry{end of stage 1}
\addplot[black!60, dashed, forget plot] coordinates {(0.25,11.1) (4,11.1)};
\node[font=\scriptsize, text=black!60, anchor=north east] at (axis cs:4,11.1) {chance};
\nextgroupplot[title={OLMo 3 7B}]
\addplot[black!45, line width=1pt, mark=*, mark size=1.8pt, error bars/.cd, y dir=both, y explicit] coordinates {(0.25,11.8) +- (0,0.2) (0.5,13.1) +- (0,0.5) (1,19.8) +- (0,2.2) (2,27.6) +- (0,3.4) (4,36.0) +- (0,3.8)};
\addplot[probecolor!55, line width=1pt, mark=square*, mark size=1.8pt, error bars/.cd, y dir=both, y explicit] coordinates {(0.25,13.6) +- (0,1.4) (0.5,22.9) +- (0,2.6) (1,61.2) +- (0,3.5) (2,95.6) +- (0,0.7) (4,99.1) +- (0,0.6)};
\addplot[probecolor, line width=1pt, mark=triangle*, mark size=1.8pt, error bars/.cd, y dir=both, y explicit] coordinates {(0.25,22.8) +- (0,1.8) (0.5,40.5) +- (0,1.6) (1,69.7) +- (0,1.2) (2,98.6) +- (0,0.8) (4,95.5) +- (0,0.3)};
\addplot[black!60, dashed, forget plot] coordinates {(0.25,11.1) (4,11.1)};
\end{groupplot}
\path (group c1r1.north west) -- (group c2r1.north east) node[midway, above=0.5cm] {\pgfplotslegendfromname{strengthsteplegend}};
\end{tikzpicture}

%% file: sections/results_details.tex
\section{No-prompt experimental details and controls}
\label{app:noprompt-details}

We ran a prespecified experiment with readouts strictly after every injected span, no instructions or question, and a fixed primary readout layer.
Fresh Gaussian directions were injected at layer 6 into one of nine numbered sentences; layer 24 activations were read at markers following the list.
We used the pretrained base model and three networks with linear and embedding weights drawn from a zero-mean Gaussian with standard deviation 0.02, biases set to zero and normalization weights set to one.
This construction retains the architecture while replacing learned weights under a specified initialization scheme; it is not OLMo's released step-0 checkpoint. The prompted sweep tests that checkpoint, but the no-prompt experiment does not.
For each seed, probes were fitted on 360 trials and evaluated on 180 balanced held-out trials, using disjoint training/test sentence pools and fresh directions.
Data seeds are 311--313 for both network states; the corresponding random-weight seeds are 411--413.
Strengths are $\alpha \in \{0, 0.25, 0.5, 1, 2, 4, 8\}$, read immediately after the list and after 16, 64 and 256 neutral suffix tokens.
Each setting has its own probe (logistic regression, $C = 0.5$, at most 2{,}000 iterations, features standardized on the training trials).
Unlike trial-level cross-validation, this design tests transfer to unseen sentences within each split; sentence pools recur across seeds.
Injection strength is a multiple of each network's training-calibrated median residual norm, which does not guarantee equal functional disruption across networks.

\begin{table}[ht]
  \centering
  \begin{tabular}{rcccc}
    \toprule
    & \multicolumn{4}{c}{Location (\%)} \\
    Strength & Gap 0 & Gap 16 & Gap 64 & Gap 256 \\
    \midrule
    \input{tables/noprompt_random.tex} 
    \bottomrule
  \end{tabular}
  \caption{No-prompt localization in randomly initialized networks at layer 24, averaged over three seeds (chance: one in nine). Gap counts neutral suffix tokens after the list. Each cell uses a separately fitted probe tested on 180 held-out trials per seed. Strength zero is an uninjected control with assigned target labels.}
  \label{tab:noprompt}
\end{table}

All prespecified random-network location tests at multipliers 4 and 8, gaps 0 and 256, and layers 24 and 58 pass 199-training-label permutation tests with Holm correction within each seed's eight-test family.
In random networks, the norm-only baseline is near chance in the primary high-strength cells.

\subsection{Pretrained comparison and uninjected controls}
The pretrained no-prompt results are retained in Table~\ref{tab:noprompt-pretrained}. Near the list, injected accuracy exceeds the pretrained model's own clean baseline. At gap 256, accuracy at strengths 4 and 8 is close to that baseline and below the random-network scores in Table~\ref{tab:noprompt}. The elevated pretrained clean scores prevent interpreting its remaining above-chance accuracy as an injection signal.

\begin{table}[ht]
  \centering
  \begin{tabular}{rcc}
    \toprule
    Strength & Gap 0 & Gap 256 \\
    \midrule
    \input{tables/noprompt_pretrained.tex} 
    \bottomrule
  \end{tabular}
  \caption{Pretrained no-prompt localization (\%). Strength-zero accuracy is elevated and remains unexplained.}
  \label{tab:noprompt-pretrained}
\end{table}

Targets are balanced and shuffled before passages are sampled; neither the target label nor its distance is included in the probe features. At zero strength, the collection hook does not modify activations. Nevertheless, the pretrained clean scores are 77/540 and 76/540 correct at gaps 0 and 256. A post-hoc check with 50{,}000 within-seed permutations of held-out labels, keeping fitted predictions fixed, gives one-sided unadjusted $p=0.014$ and $p=0.018$. These are correlated checks on the same trials, not independent replications, and do not include refitting or adjustment for the broader set of examined cells. The corresponding random-network checks give $p=0.411$ and $p=0.580$. We have not established the cause of the pretrained elevation, so we compare injected scores with the observed clean baseline as well as nominal chance.

An additional post-hoc diagnostic refitted the pretrained clean probes with 199 permutations of training labels per seed and gap. Comparing with original-label refits in the same local environment gives two-gap Holm-adjusted $p=0.010$ and $p=0.035$. The local refits change the near-gap aggregate by one correct trial and leave the long-gap aggregate unchanged; the table retains the original scores. These checks do not resolve the anomaly.

\subsection{Transfer to new text and formatting}

For these transfer tests, the linear probe selects $C\in\{0.1,0.5,2\}$ on a stratified 270/90 split of the original training trials, then refits on all 360 training trials before being frozen.
Frozen linear probes transfer to fresh numbered text at \npFreshLo--\npFreshHi\% in random networks at multiplier 8 and gap 0, but fall to \npUnlabelledLo--\npUnlabelledHi\% when numbering is removed from the same text.
Removing sentence labels also shortens the input, changing token positions and the distance to the readout.

Thus successful decoding within a distribution does not imply generalization across text and formatting; failed frozen transfer also does not establish that the representation contains no recoverable information.

\subsection{Propagation control and numerical replay}
\label{app:path-details}

Paired injected-minus-clean activations establish propagation independently of classifier accuracy.
Across both network states and all three seeds, blocking all outward attention from the injected span after layer 6 eliminates the perturbation at later readouts exactly.
Injecting after the capture layer also leaves captured features unchanged.
Numerical replay is imperfect: default attention does not always reproduce stored activations within the prespecified tolerance across processes.
With explicit math attention, 18 audited trials per network state reproduce exactly across two processes, but still differ from the original stored activations. Across the full 180-trial original-input replay for all six networks/seeds, the frozen-decoder checks change accuracy by at most 2.22 percentage points and predictions by at most 5\%. These checks use the transfer experiment's frozen decoders and do not constitute a refitted replication of every cell in Table~\ref{tab:noprompt}.

%% file: tables/noprompt_random.tex
0 & 11.5 & 11.3 & 11.9 & 10.9 \\
1 & 12.2 & 13.1 & 13.3 & 11.5 \\
2 & 16.9 & 18.1 & 15.9 & 13.5 \\
4 & 27.4 & 26.9 & 21.9 & 19.1 \\
8 & 31.5 & 29.8 & 23.5 & 21.5 \\

%% file: tables/noprompt_pretrained.tex
0 & 14.3 & 14.1 \\
1 & 24.3 & 13.5 \\
2 & 50.7 & 14.1 \\
4 & 53.0 & 14.8 \\
8 & 48.1 & 14.3 \\